\documentclass[final]{cvpr}
\usepackage{times}
\usepackage{amsmath}
\usepackage{amssymb}
\usepackage{booktabs}
\usepackage{chngcntr}
\usepackage{cuted}
\usepackage{graphicx}
\usepackage{multirow}
\usepackage{nicefrac}
\usepackage{pifont}
\usepackage{subcaption}
\usepackage{tabularx}
\usepackage{url}
\usepackage{xcolor}
\usepackage{xspace}
\usepackage[pagebackref=true,breaklinks=true,colorlinks,bookmarks=false]{hyperref}
\usepackage{cleveref}
\usepackage{pifont}
\usepackage[symbol]{footmisc}
\usepackage[shortlabels]{enumitem}

\def\cvprPaperID{555}
\def\confYear{CVPR 2021}

\newcommand{\ignoreme}[1]{}
\newcommand{\cmark}{\ding{51}}%
\newcommand{\xmark}{\ding{55}}%

\makeatletter
\renewcommand{\paragraph}{%
  \@startsection{paragraph}{4}%
  {\z@}{0.5em}{-1em}%
  {\normalfont\normalsize\bfseries}%
}
\makeatother

\title{Discovering Relationships between Object Categories \\via Universal Canonical Maps}
\author{Natalia Neverova*, Artsiom Sanakoyeu*, Patrick Labatut, David Novotny, Andrea Vedaldi\\
Facebook AI Research\\
{\tt\small \{nneverova,asanakoy,dnovotny,plabatut,vedaldi\}@fb.com}
}

\begin{document}
\twocolumn[\maketitle\vspace{-1.7em}]
\footnotetext[1]{Both authors contributed equally to this work.}

\begin{strip}
\vspace{-3em}
\centering
\includegraphics[width=\linewidth]{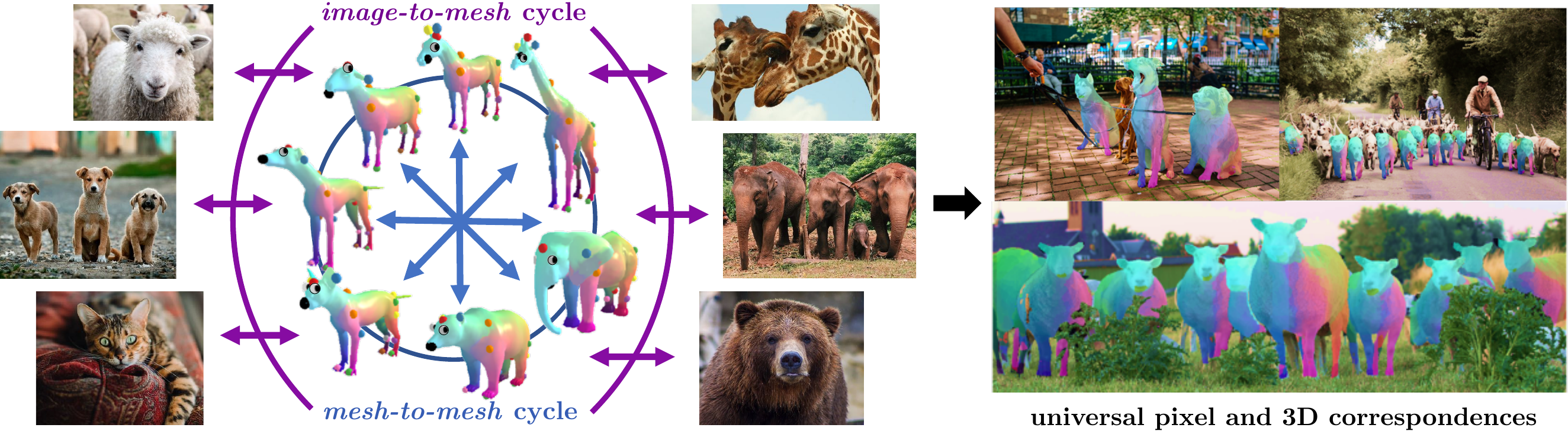}\vspace*{5pt}
\captionof{figure}{\textbf{Inter-category correspondences emerge from dense pose prediction.}
Our method discovers high-quality correspondences between different object classes automatically, as a byproduct of learning category-specific dense pose predictors.
It does so by enforcing cycle consistency between reference 3D templates as well as by a new type of consistency between images and templates.
This allows the model to transfer information between animal classes (\eg the location of the eyes).
}\label{fig:splash}
\vspace*{5pt}
\end{strip}

\begin{center}
  \large\textbf{Abstract}
\end{center}

\emph{We tackle the problem of learning the geometry of multiple categories of deformable objects jointly.
Recent work has shown that it is possible to learn a unified dense pose predictor for several categories of related objects.
However, training such models requires to initialize inter-category correspondences by hand.
This is suboptimal and the resulting models fail to maintain correct correspondences as individual categories are learned.
In this paper, we show that improved correspondences can be learned automatically as a natural byproduct of learning category-specific dense pose predictors.
To do this, we express correspondences between different categories and between images and categories using a unified embedding.
Then, we use the latter to enforce two constraints: symmetric inter-category cycle consistency and a new asymmetric image-to-category cycle consistency.
Without any manual annotations for the inter-category correspondences, we obtain state-of-the-art alignment results, outperforming dedicated methods for matching 3D shapes.
Moreover, the new model is also better at the task of dense pose prediction than prior work.}
\section{Introduction}\label{s:intro}

Algorithms can nowadays understand well the geometry of \emph{specific object categories} such as humans:
we have reliable methods for detecting and segmenting them, extracting their 2D landmarks and dense surface coordinates, as well as reconstructing them in 3D.
In principle, these methods can be applied to many other types of objects, such as any kind of animal, from pets to wildlife.
In practice, however, doing so is often 
prohibitively expensive.
The main bottleneck is data acquisition, especially for supervised training in 3D, and extensive manual annotation.
High-quality 3D human models are bootstrapped using specialized motion capture systems such as domes that are difficult to apply to objects such as wild animals.
Annotating 2D geometric primitives such as segments and 2D keypoints can be done manually from raw images, but it is costly and somewhat difficult to do for unfamiliar animal anatomies.
Thus, a na\"{\i}ve application of existing high-quality model acquisition techniques cannot trivially scale to learning the massive variety of object types that exist in the world, which include 6.5K mammal species, 7.7M animal species, and around 8.7M natural species overall~\cite{burgin18how-many,mora11how-many}.

The key to scaling is to realize that, while there are indeed millions of different types of objects, these are not independent.
For instance, different cat breeds are relatively similar, so a single `cat' model is likely to work well for all cats, just like a single `human' model has been shown to work well for many different human body shapes~\cite{loper15smpl:}.
In fact, useful information can likely be shared among fairly different types of objects, such as all mammals or all animals.
The limit is given by the ability of the model to represent diverse information while capturing and eliminating redundancies wherever possible.
The hope is that such a model could learn the geometry of different object types with a cost which is sub-linear in their number.

A similar idea was recently pursued in~\cite{neverova20continuous} for the task of \emph{dense pose prediction}~\cite{guler18densepose:}.
Just like 2D pose prediction estimates the location of a small number of distinctive object landmarks, dense pose estimation does so for a continuous set of landmarks, identified as the point of a 3D template of the object
(\cref{fig:splash}).
The goal is to learn a \emph{canonical map}, i.e.~a function that maps all relevant pixels in an image to the corresponding points in the template, thus identifying them.
For supervised learning, correspondences between images and templates are collected manually, using a category-specific template for each example object.
As a result, annotations for different object categories are unrelated, which makes it hard to learn a universal, category-agnostic object representation.

In order to address this problem, the authors of~\cite{neverova20continuous} establish initial point-to-point correspondences between different category-specific templates using a mix of manual annotations and automated interpolation.
However, as we show in the experiments, their approach has two shortcomings.
Firstly, their manual correspondence initialization is somewhat arbitrary and thus likely suboptimal.
The second problem, which partially arises from the first, is that their initial inter-category correspondences are not \emph{maintained} while the model is trained, and are eventually `forgotten'.

In this paper we argue that, if the goal of the alignment is to facilitate learning a multi-category object representation, an optimal alignment should emerge spontaneously as part of the learning process, thus solving the two issues above.
Our key contribution is thus a new learning formulation for universal canonical maps that induces \emph{automatically} high-quality intra-category correspondences.
The most important outcome is that the learned maps solve the dense pose prediction problem accurately for several object categories while at the same time putting those in correspondence, allowing to transfer information between them.

We base our model on learning a single, universal embedding space to express all required correspondences.
Points in the different 3D templates as well as image pixels are mapped to this common space, which allows to compute dense template-to-template and image-to-template correspondences.
Differently from~\cite{neverova20continuous}, the template embeddings in this work are \emph{not} initialized from manually annotated inter-category correspondences.
Instead, all embeddings are obtained automatically while learning the canonical maps for individual categories while satisfying certain consistency constraints.


For the constraints, we use simple but effective rules.
Apart from the most basic one, which encourages similarity of the embeddings of nearby template points (smoothness), we contribute by introducing two types of cycle-consistency for learning canonical surface mappings:
The first one enforces cycle consistency between different 3D templates, which encourages bijective correspondences between them.
Additionally, we note that canonical maps, by establishing correspondences from images to templates, are not bijective but injective, and we show that this can be exploited by an asymmetric form of cycle consistency between images and templates.
By using the common embedding space, all such constraints are expressed as differentiable loss terms.

Empirically, we demonstrate several advantages of our new approach compared to~\cite{neverova20continuous}.
We show that our approach finds automatically high-quality correspondences between different object categories \emph{without any manual supervision for this task}.
This is compelling because it shows that, as we hypothesized, there is a natural advantage in learning jointly the geometry of different but related object types.
In fact, the 3D correspondences we discover in this manner outperform the ones discovered by state-of-the-art 3D shape alignment methods.
Finally, our method not only aligns canonical maps, but also improves their quality, resulting in more accurate dense pose prediction than the state of the art.


\section{Related work}\label{s:related}

\paragraph{Human pose estimation.}

Human pose prediction often starts by detecting 2D landmarks, usually coinciding with the main joints of the body~\cite{lin14microsoft,andriluka142d-human,johnson11learning,johnson10clustered}.
For this task, early shallow methods~\cite{felzenszwalb08a-discriminatively,bourdev09poselets:,johnson11learning,ramanan06learning} have been surpassed by deep convolutional architectures~\cite{newell16stacked,wei16convolutional,cao17realtime}.
Sparse landmarks can be replaced by dense ones, identifiable with a reference 3D template of the object.
The resulting dense pose prediction problem was pioneered by 
DensePose~\cite{guler18densepose:} using the SMPL~\cite{loper15smpl:} mesh as a canonical template.
Parsing R-CNN~\cite{yang2019parsing} improved the Dense Pose network by extending the popular R-CNN architecture~\cite{girshick13rich}.
More recently, Slim DensePose~\cite{neverova19slim} showed that a smaller number of keypoint annotations is sufficient to learn competitive DensePose models, potentially significantly reducing the effort required for learning new non-human categories.

\paragraph{Animal pose estimation.}

Several works also attempted to estimate the pose of various animal species.
Methods such as \cite{zhang14part-based,singh2016learning,kanazawa16warpnet:,kanazawa18learning,novotny19c3dpo,welinder10caltech-ucsd} learned to detect~\cite{zhang14part-based,singh2016learning}, match~\cite{kanazawa16warpnet:} or reconstruct \cite{kanazawa18learning,novotny19c3dpo} various birds from the CUB dataset \cite{welinder10caltech-ucsd}. 
3D reconstruction of the shape of a broader set of animal species has been attempted by Zuffi et al. in \cite{zuffi2017menagerie,zuffi2018lions,zuffi2019safari}.
Similar to monocular 3D human mesh recovery models \cite{kanazawa18end-to-end,kolotouros19convolutional,kolotouros19learning} that predict parameters of SMPL, \cite{zuffi2017menagerie,zuffi2018lions,zuffi2019safari} utilize a parametric model of a mesh of an animal body (SMAL) in order to constrain the set of possible animal reconstructions, with further improvements in the work of Biggs et al.~\cite{biggs2018,biggs2020left}. Sanakoyeu et al.~\cite{sanakoyeu20transferring} transfer DensePose from humans to proximal animal classes without extra labels by a self-training approach.

The work most relevant to ours is Neverova et al.~\cite{neverova20continuous}, which introduced the idea of continuous surface embeddings (CSE) to tackle the dense pose prediction problem for several animal categories together.
They further contributed a dataset of dense pose annotations for various animal species.
We improve on~\cite{neverova20continuous} by learning more accurate canonical maps that are more consistent across different categories and by not requiring any manual initialization for the correspondences between different object categories.
We also contribute with an extended dataset of dense animal poses for experimentation.

\paragraph{Intrinsic 3D shape analysis.}

Our work is also related to the analysis of the intrinsic properties of 3D shapes, where the fundamental problem is to establish correspondences between different shapes.
Non-deep learning methods include embeddings of geodesic distance matrices \cite{elad2003bending,bronstein2006generalized} and various kinds of diffusion geometry~\cite{coifman2005diffusion} descriptors --- Heat Kernel Signature~\cite{sun2009concise} and its scale-invariant follow-up~\cite{bronstein2010scale}, Gromov-Hausdorff descriptors~\cite{bronstein2010gromov} and the Wave Kernel Signature~\cite{aubry2011wave}.
One of the main building blocks of the aforementioned descriptors are the eigenfunctions of the Laplace-Beltrami operator (LBO) \cite{rustamov2007laplace} that define a smooth basis of a coordinate frame of a mesh surface.
Ovsjanikov et al. ~\cite{ovsjanikov12functional} proposed the functional map (FM) framework that establishes soft correspondences between pairs of shapes by relating the mesh LBO eigenfunctions with a simple linear mapping.
The CSE method from~\cite{neverova20continuous} uses FMs and ZoomOut~\cite{melzi2019zoomout} to interpolate an initial set of manually-established inter-class correspondences.
Differently from them, we only use LBO to express smoothness, but we otherwise consider topological constraints such as bijectivity and injectivity that are more appropriate for establishing non-isometric correspondences, such as between different animal categories.

\paragraph{Cycle consistency.}

Cycle consistency is a powerful paradigm that has been explored in many different fields of computer vision: pixel-wise image matching~\cite{zhou2015multi,zhou15flowweb:}, image translation~\cite{zhu17unpaired}, or category-specific 3D reconstruction~\cite{zhou2016learning}.
Given a single input image of an instance of an object category, Kulkarni et al.~\cite{kulkarni19canonical,kulkarni2020articulation} enforce consistency between a rendered UV map of a 3D template shape of the object category and the learned canonical map, while our method does not require to fit/render the 3D model.
In the context of 3D shape analysis, Huang et al.~\cite{huang2013consistent} introduced a semi-definite programming formulation that factorized a matrix of all point-to-point matches between pairs of meshes in order to make the matches cycle-consistent. 
Similarly, Yang et al.~\cite{yang20mapping} use the Sinkhorn regularization (SH) to find the nearest cycle-consistent solution to an initial matrix of noisy point-wise matches. Ren et al.~\cite{ren2020maptree} exploit the spectral properties of correspondences and cycle consistence between shape pairs.
Our method is inspired by~\cite{yang20mapping} in the sense that we utilize cycle consistency in order to improve our dense pose labels by relating surfaces of different category template shapes.

\section{Method}\label{s:method}

We start by summarizing the continuous surface embedding (CSE) representation of~\cite{neverova20continuous} (\cref{s:cse}) and then we explain how to extend it to learn high-quality inter-category correspondences automatically (\cref{s:correspondences}).

\subsection{Continuos surface embeddings}\label{s:cse}

The \emph{continuous surface embedding} (CSE) of~\cite{neverova20continuous} allows us to express correspondences between different 3D templates and between templates and images in a homogeneous and differentiable manner.
A CSE is a function $e : S \rightarrow \mathbb{R}^D$ sending each point $X \in S$ of a 3D surface~$S$ to a $D$-dimensional embedding vector.
We assume that the surface $S$ is a mesh with $K$ vertices
$
S = (X_k)_{1\leq k \leq K}
$
and we collect the corresponding embedding vectors as the rows of a matrix $E \in \mathbb{R}^{K\times D}$.
The matrix $E$, which we learn from data, can be fairly large, but
smoothness\footnote[2]{\Ie the fact that nearby vertices should have similar embeddings.} can help to reduce its dimensionality.
This can be done by considering a smooth functional basis $U \in \mathbb{R}^{K \times Q}$ on the mesh, where $Q \ll D$, and define $E = U \hat E$.
With this, we can work with the compressed embedding parameterization $\hat E \in \mathbb{R}^{Q \times D}$.
As in~\cite{neverova20continuous}, we take $U$ to be the lowest eigenvectors of the Laplace-Beltrami operator (LBO) of the mesh $S$.
While the LBO is often used in the literature as a cue to match near-isometric shapes, our shapes are not at all isometric.
For this reason, we use the LBO only to encode a generic notion of smoothness, but \emph{not} as a cue for matching.


\paragraph{Encoding correspondences via CSEs.}


Embedding vectors can be used to define correspondences between any two sets of objects $A = (a_1,\dots,a_K)$ and $B=(b_1,\dots,b_L)$.
Namely, given embedding functions $e : A \rightarrow \mathbb{R}^D$ and $e : B \rightarrow \mathbb{R}^D$, we can compare embedding vectors to send elements of set~$B$ to elements of set~$A$ probabilistically:
\begin{equation}\label{e:general}
p(a_k | b_l, e)
=
\frac{
  \exp\left(
  -\langle e_{a_k}, e_{b_l} \rangle
  \right)
}{
  \sum_{t=1}^K \exp\left(
  -\langle e_{a_t}, e_{b_l} \rangle
  \right)
}.
\end{equation}

In our case, given two CSEs $E=U\hat E$ and $E' = U' \hat E'$ for two different meshes $S$ and $S'$, \cref{e:general} gives us distributions $p(X_k | X'_l, \hat E, \hat E')$ and $p(X'_l | X_k, \hat E, \hat E')$, encoding mappings $S' \rightarrow S$ and $S \rightarrow S'$, respectively.
%
We can also express image-to-mesh and mesh-to-image maps.
For this, let $I$ be an image and consider a finite set $\Omega \subset \mathbb{R}^2$ of pixel locations.
We use a deep convolutional neural network $e_x = [\Phi(I)]_x$ to compute the embedding vectors for all the pixels $x \in \Omega$.
Then, given a mesh $S$ together with its embedding matrix $E = U \hat E$, we can use~\cref{e:general} to obtain a distribution $p(X_k | x, \hat E, \Phi(I))$ encoding a map $\Omega \rightarrow S$ from the pixels to the mesh.
Note that the latter is, by definition, a canonical map, and as such it provides a solution to the dense pose prediction task.
We can also swap the roles of image and mesh in this expression, obtaining a probability $p(x | X_k, \hat E, \Phi(I))$ encoding a reverse map $S \rightarrow \Omega$.
This map will be useful later.

Finally, we can, in an entirely analogous manner, define  image-to-image correspondences $\Omega\rightarrow\Omega'$ by comparing embeddings $\Phi_x(I)$ and $\Phi(I')_{x'}$.
This is useful to transfer information directly across images, as we demonstrate in the experiments for keypoint transfer.


\paragraph{Working with multiple object categories.}

The approach above can easily accommodate any number of categories and corresponding templates.
Each category $m=1,\dots,M$ is captured by a mesh and its embedding $(S^m, \hat E^m)$.
Each mesh can have a different number of vertices $|S^m| = K^m$.
Likewise, the LBO basis $U^m \in \mathbb{R}^{Q^m \times D}$ is mesh-specific, including having potentially a different number of basis elements $Q^m$.
Crucially, however, the dimensionality of the embedding space $D$ is the same for all templates, as the embeddings must be comparable.


\subsection{Dense pose and emerging correspondences}\label{s:correspondences}

The use of a common embedding space for templates and images means that all such objects can be put in correspondence by using the method of~\cref{s:cse}.
However, this does not necessarily mean that the correspondences learned by the model are meaningful.
In more detail, manual annotations for the dense pose task are of the type $(I, m, x, X)$ where $I$ is an image, $m$ a category, $x$ a pixel, and $X \in S^m$ its corresponding vertex in the category-specific template $S^m$~\cite{guler18densepose:,neverova20continuous}.
By fitting such annotations, the model is encouraged to learn good dense pose predictors for each category, but not necessarily good inter-category correspondences.
%
The latter may emerge because the neural network~$\Phi$ is shared in full or in part among different categories, which means that similarly-looking images will naturally tend to be embedded in similar ways.
However, this is a weak effect.
Below, we add several constraints to improve the quality of the emerging correspondences.

\paragraph{Dense pose supervision.}

Solving the dense pose prediction tasks means learning maps $\Omega \rightarrow S^m$ sending the image region $\Omega$ that contains an occurrence of the object to the template $S^m$ of the object itself.
As noted above, supervision for this task comes in the form of a dataset $\mathcal{D}$ of tuples $(I, m, x, X)$ and is captured by the loss as follows:
\begin{multline}\label{e:loss-super}
\mathcal{L}^\text{sup}
=
\frac{1}{|\mathcal{D}|}
\sum_{(I,m,x,X)\in\mathcal{D}}
\sum_{k=1}^{K^m}
d_{S^m}(X_k^m, X) \\ 
\cdot p(X_k^m | x, \hat E^m, \Phi^m(I)).
\end{multline}
In this expression, $d_{S^m}$ is the geodesic distance on the mesh $S^m$.
This loss is optimized w.r.t.~the mesh embeddings and neural networks $(\hat E^m, \Phi^m)_{1\leq m \leq M}$ (where the different networks share most or all of their parameters).

\paragraph{Inter-category correspondences.}\label{s:mesh-to-mesh}

We assume that there exists sensible one-to-one correspondences $S^m \leftrightarrow S^n$ between any pair of templates.
In this case, the cycle $S^m \rightarrow S^n \rightarrow S^m$ should approximate the identity function.
We can rewrite the cycle in terms of the probabilistic correspondences described in~\cref{s:cse} by marginalizing the intermediate step as follows:
\newcommand{\no}[1]{}
\begin{equation}\label{e:p-auto}
  p(X^m_k | X^m_t\no{, \hat E^m, \hat E^n})
   =
   \sum_{l=1}^{K^n}
   p(X^m_k | X^n_l\no{, \hat E^m, \hat E^n})
   \,
   p(X^n_l | X^m_t\no{, \hat E^m, \hat E^n}).
\end{equation}
While we do not show it for compactness, note that all such probabilities depend on the learned embeddings $\hat E^m$ and $\hat E^n$.
If the cycle is closed correctly, this probability should peak at $X^m_t = X^m_k$, which is captured by the \emph{mess-to-mesh} loss (\textbf{m2m}):
\begin{equation}\label{e:loss-m2m}
  \mathcal{L}^{mn}
  =
  \frac{1}{K^m}\sum_{k=1}^{K^m}
  \sum_{t=1}^{K^m}
  d_{S^m}(X^m_k, X^m_t)\, p(X^m_k | X^m_t\no{, \hat E^m, \hat E^n}).
\end{equation}
To the loss $\mathcal{L}^{mn}$ we also add the symmetric loss $\mathcal{L}^{nm}$.
Cycle consistency has been exploited before in many different contexts~\cite{kulkarni19canonical,wang19learning,yang20mapping,zhu17unpaired,zhou16blearning,huang13consistent,huber02automatic}.
Here we use it to guide the discovery of correspondences between different meshes.



\paragraph{Canonical map injectivity.}

The signal~\eqref{e:loss-super} is only given at a sparse set of manually-labelled image pixels.
A denser constraint can be obtained by noting that the canonical maps $\Omega \rightarrow S^m$ must be \emph{injective}, in the sense that all pixels in the object region $\Omega$ should map to different vertices in the mesh $S^m$.
Injectivity means that the canonical map has a left inverse:
if a mesh vertex corresponds to at least one image pixel, then this correspondence must be unique.
We can thus close the cycle $\Omega \rightarrow S^m \rightarrow \Omega$, resulting in the \emph{image-to-mesh} loss (\textbf{i2m}):
\begin{equation}\label{e:loss-i2m}
    \mathcal{L}^{Im}
    =
    \frac{1}{|\Omega|}
    \sum_{x\in \Omega}\,
    \sum_{y\in \Omega}
    d_{I}(y, x) p(y | x\no{,\hat E^m, \Phi^m(I)}).
\end{equation}
where $d_I$ is a distance in image space (e.g.~Euclidean) and, similar to~\cref{e:p-auto},
\begin{equation}\label{e:p-auto-im}
  p(y|x\no{,\hat E^m, \Phi^m(I)})
  =
  \frac{1}{K^m}\sum_{k=1}^{K^m}
  p(y | X^m_k\no{,\hat E^m, \Phi^m(I)})\\
  \, p(X^m_k | x\no{,\hat E^m, \Phi^m(I)}).
\end{equation}
While not shown for compactness, all these probabilities depend on the mesh embedding $E^m$ and the neural network $\Phi^m$ that we wish to learn.

Rather than summing~\cref{e:loss-i2m} on the entire set $\Omega$, we consider a \emph{downsampled} version $\bar \Omega \subset \Omega$ with $|\bar\Omega|\ll |\Omega|$.
This is done for computational efficiency (as there can be a very large number of pixels in certain image regions).
Compared to using the full domain, the effect is to slightly relax~\cref{e:loss-i2m}.

Note that, differently from the mesh-to-mesh cycle, this cycle is \emph{not} symmetric:
while we can close the chain $\Omega \rightarrow S^m \rightarrow \Omega$, we \emph{cannot} close the chain $S^m \rightarrow \Omega \rightarrow S^m$.
The first chain is valid because all pixels in $\Omega$ correspond to \emph{a unique point} of the mesh $S^m$.
On the other hand, many of the points in the mesh $S^m$ will \emph{not} have a corresponding image point in $\Omega$ for the simple fact that at least part of the object cannot be visible in a given image.

\subsection{Overall loss}

To summarize, our model is trained by minimizing a combination of the losses of~\cref{e:loss-super,e:loss-m2m,e:loss-i2m}:
$$
\mathcal{L}^\text{sup} +
\frac{1}{M(M-1)}
\mathop{\sum_{m,n=1}^M}_{m\not=n} \mathcal{L}^{mn} +
\frac{1}{|\mathcal{D}|} \sum_{(I,m)\in\mathcal{D}} \mathcal{L}^{Im}.
$$

\section{Experiments}\label{s:exp}

\begin{table}[!t] 
\centering
\scalebox{0.75}{
\begin{tabular}{c|c|c|c|cc}
\toprule
  dataset & init. & \!train\! & $\textbf{AP} \downarrow$ & $\textbf{GErr} \downarrow $ & $\textbf{GPS} \uparrow $ \\
  \midrule
  \multirow{2}{*}{DP-LVIS~\cite{neverova20continuous}} &
  \multirow{2}{*}{ZoomOut} & -- & 25.4 & 23.7 & 66.7\\
  & & \checkmark & 35.1 & 28.0 & 68.7\\
  \cmidrule{2-6}
  & Random & \checkmark & 34.4 & 34.1 & 63.7\\
\midrule
  DP-LVIS v1.0 &
  Random & \checkmark & 37.4 & 20.7 & 77.1 \\
\bottomrule
\end{tabular}}
\caption{
\textbf{Baselines (humans \& animals).}
We train a universal canonical map using DensePose-COCO and animal data and report DensePose performance on animal categories (\textbf{AP}), as well as mesh alignment quality for animals and people (\textbf{GErr} and \textbf{GPS}). The architecture is of~\cite{neverova20continuous}, combined with multi-class detection. ZoomOut initialization does not result in performance gains on a larger dataset.
}\label{tab:dp80k}
\end{table}

\ignoreme{
\begin{table*}[!t] 
\centering
\scalebox{0.78}{
\begin{tabular}{cc|c|ccc|cc|ccc|cc|cc}
\toprule
  schedule & init. & \hspace*{\fill} train \hspace*{\fill} & $\textbf{AP}$ & $\textbf{AP}_{50}$ & $\textbf{AP}_{75}$ & $\textbf{AP}_{M}$ & $\textbf{AP}_{L}$ & $\textbf{AR}$ &
  $\textbf{AR}_{50}$ & $\textbf{AR}_{75}$ & $\textbf{AR}_{M}$ &
  $\textbf{AR}_{L}$ & $\textbf{GErr}$ & $\textbf{GPS}$ \\
  \midrule
  \parbox[t]{2mm}{\multirow{5}{*}{1x}} &
  \multirow{2}{*}{ZoomOut} & -- \\
  & & \checkmark & \\
  \cmidrule{2-15}
  & Random & \checkmark & 37.4 & 69.0 & 35.7 & 35.0 & 41.9 & 51.7 & 79.7 & 53.1 & 41.8 & 55.5 & 16.53 & 81.60\\
\bottomrule
\end{tabular}}
\caption{\textbf{Staring point vs NeurIPS.} DensePose performance. GPSm metric for AP/AR, trained on people + animals, tested on animals only. 80k iterations.}\label{tab:dp80k}
\end{table*}
}
\begin{figure*}[!t]
\begin{center}
\includegraphics[width=0.95\linewidth]{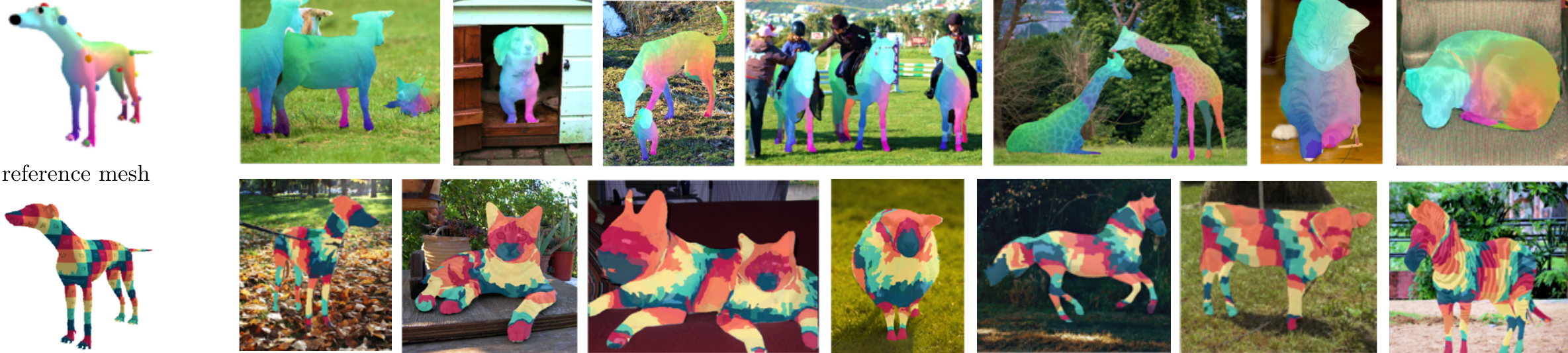}
\caption{\textbf{Qualitative results by the full model \textbf{m2m+i2m}-all.} The \texttt{dog} 3D model serves as a common reference for all classes. 
This setup is significantly more challenging than in~\cite{neverova20continuous}, where each class was visualised with its own 3D  reference.}\label{f:dpresults}
\vspace*{-10pt}
\end{center}
\end{figure*}

\begin{table*}[!t] 
\centering
\scalebox{0.92}{
\begin{tabular}{l|cc|ccc|cc|ccc|cc}
\toprule
  \hspace*{\fill} method \hspace*{\fill} &
  $\textbf{GErr} \downarrow$ & $\textbf{GPS} \uparrow$ &
  $\textbf{AP}$ & $\textbf{AP}_{50}$ & $\textbf{AP}_{75}$ & $\textbf{AP}_{M}$ & $\textbf{AP}_{L}$ & $\textbf{AR}$ &
  $\textbf{AR}_{50}$ & $\textbf{AR}_{75}$ & $\textbf{AR}_{M}$ &
  $\textbf{AR}_{L}$ \\
  \midrule
  our baseline & 14.05 & 84.33 & 37.5 & 67.8 & 36.4 & 35.1 & 41.9 & 51.5 & 78.8 & 53.2 & 41.7 & 55.1 \\
  w/ \textbf{m2m} & 11.96 & 87.35 & 38.2 & 68.5 & 36.4 & 36.6 & 42.6 & 52.0 & 79.7 & 52.6 & 42.9 & 55.6 \\
  w/ \textbf{i2m} & 12.67 & 85.13 & 38.1 & 68.7 & 36.2 & 35.5 & 42.4 & 52.0 & 79.6 & 53.2 & 42.7 & 55.6 \\
  w/ \textbf{i2m}-all & 11.74 & 87.48 & 38.3 & \textbf{68.9} & 36.3 & 35.7 & 42.5 & 52.3 & \textbf{79.9} & 53.5 & 42.8 & 55.7 \\
  w/ \textbf{m2m+i2m} & 11.37 & 88.14 & 38.3 & 68.7 & 36.6 & 36.7 & 42.5 & 52.3 & 79.7 & 53.7 & 43.0 & 55.7\\
  w/ \textbf{m2m+i2m}-all & \textbf{10.90} & \textbf{88.85} & \textbf{38.5} & 68.7 & \textbf{37.1} & \textbf{37.5} & \textbf{42.6} & \textbf{52.5} & 79.7 & \textbf{54.3} & \textbf{43.8} & \textbf{55.9}\\
\bottomrule
\end{tabular}}
\caption{\textbf{Ablation of cycle-consistency loss terms (animals only):} \textbf{i2m} corresponds to comparing the image to the target mesh given the ground truth class label, while \textbf{i2m}-all matches all object instances to all meshes in a cross-category regime.\label{tab:dpresults}}
\end{table*}

\begin{figure}[!t]
\begin{center}
\vspace{-10pt}  
\includegraphics[width=0.95\linewidth]{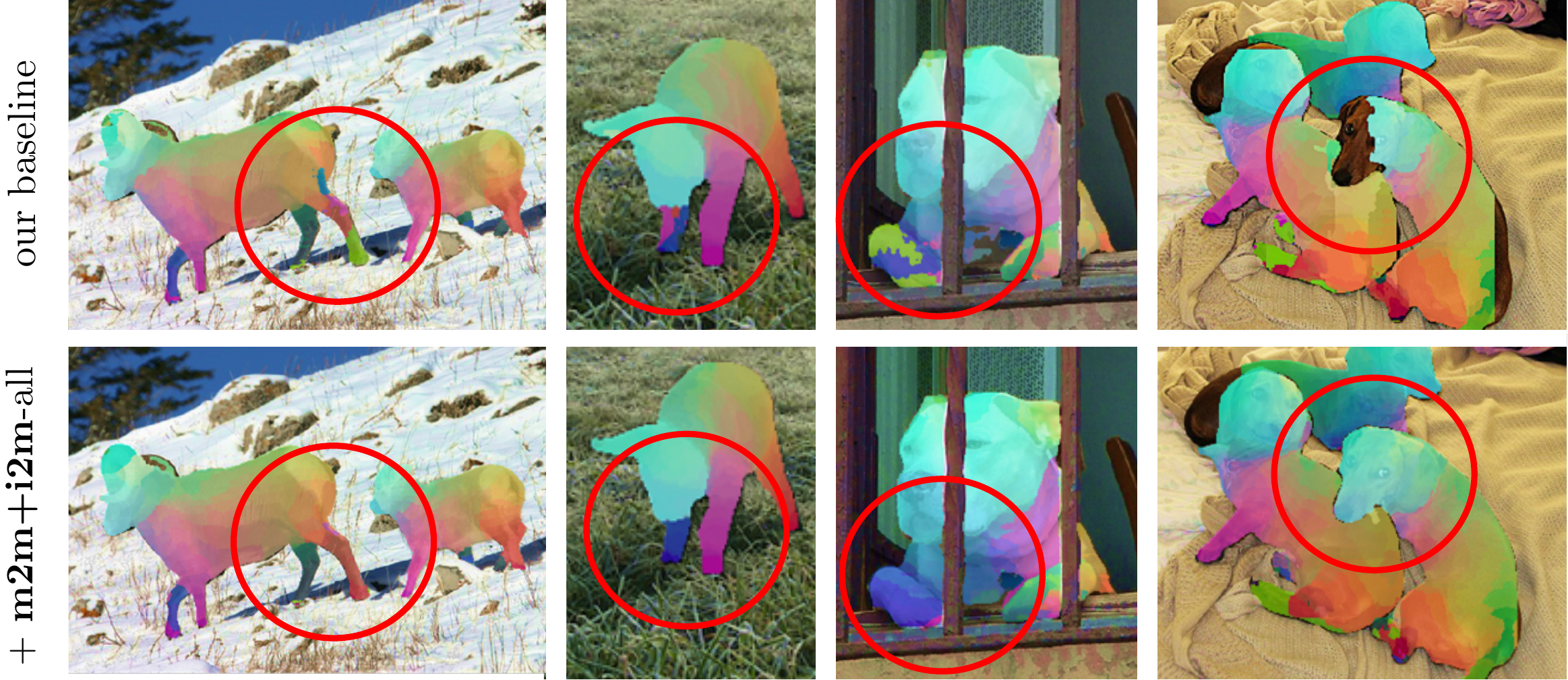}
\caption{\textbf{Effect of the \textbf{m2m+i2m}-all term:} improved local consistency and smoothness in dense pose predictions (see outlined regions),
as well as more accurate 
instance 
masks.\vspace*{-15pt}}\label{f:dpresults_i2m}
\end{center}
\end{figure}
\begin{figure*}[!t]
\centering
\begin{minipage}{0.35\linewidth}
\centering
\vspace*{5pt}
\includegraphics[width=0.62\linewidth]{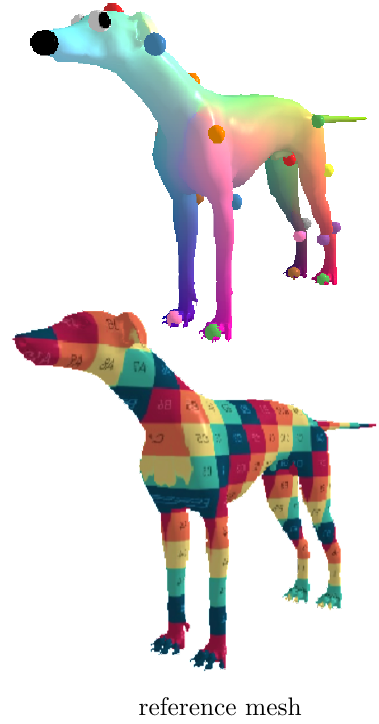}\vspace*{7pt}\\
\scalebox{0.8}{
\begin{tabular}{l|cc}
\toprule
  \hspace*{\fill} method \hspace*{\fill} &
  \!$\textbf{GErr} \downarrow$\! & \!$\textbf{GPS}\! \uparrow$ \\
  \midrule
  ZoomOut~\cite{melzi2019zoomout} & 26.24 & 63.33 \\
  Deep Shells~\cite{eisenberger2020deep} & 17.91 & 78.68 \\
  \midrule
  our baseline & 14.05 & 84.33 \\
  + \textbf{m2m+i2m}-all (best $\mathbf{AP}$) & 10.90 & 88.85 \\
  + \textbf{m2m+i2m}-all* (best $\mathbf{GErr}$)\!\! & \textbf{9.42} & \textbf{91.14}\\ 
\bottomrule
\end{tabular}}
\end{minipage}
\begin{minipage}{0.63\linewidth}
\centering
\vspace{-10pt}
\includegraphics[width=0.97\linewidth]{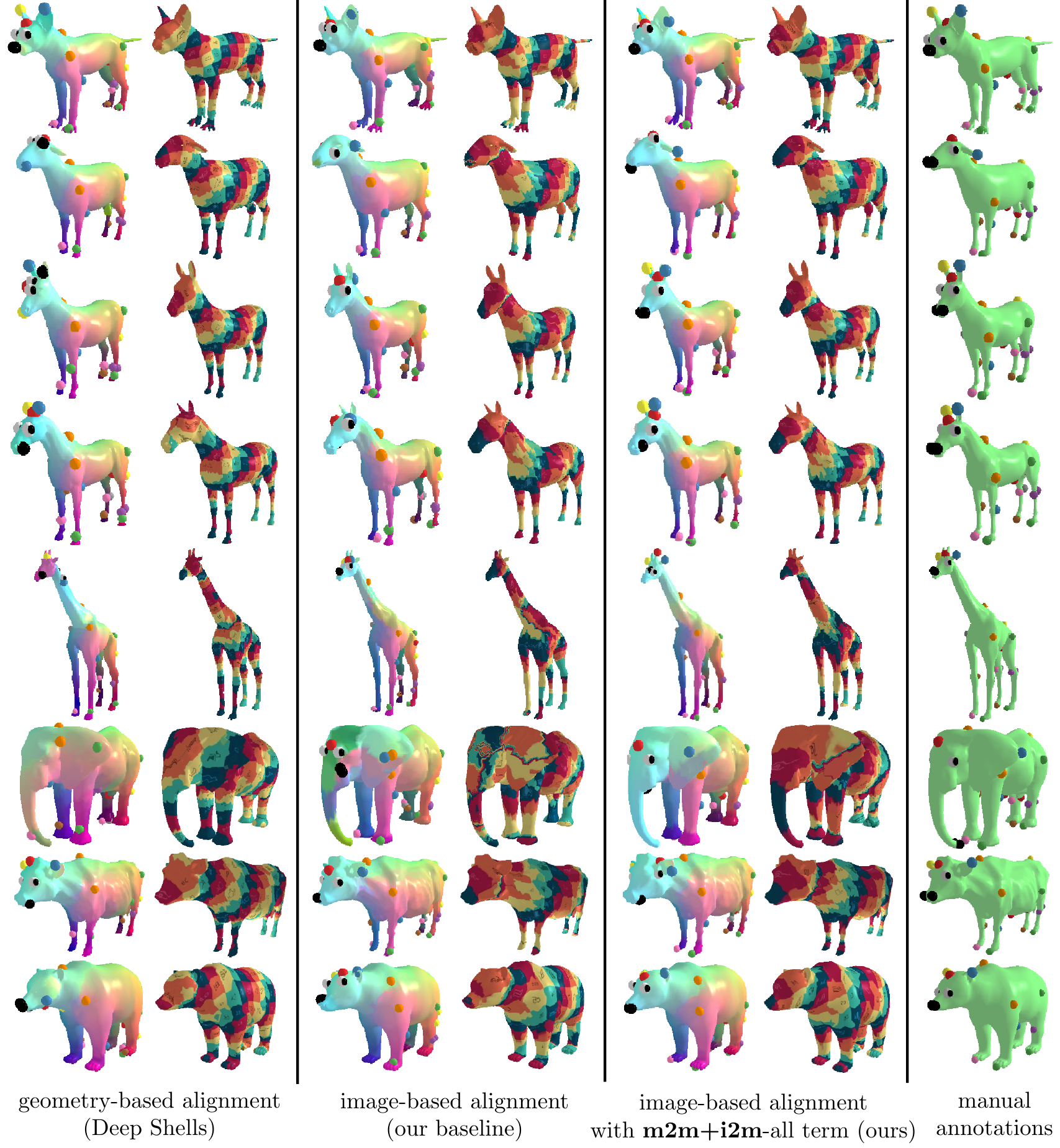}%
\end{minipage}
\caption{\textbf{3D mesh alignment: 9 animals.} *number obtained with a 10x increased weight of the \textbf{m2m+i2m}-all term.\label{fig:3Dalign}} 
\end{figure*}

\begin{figure*}[!t]
\centering
\includegraphics[width=0.95\linewidth]{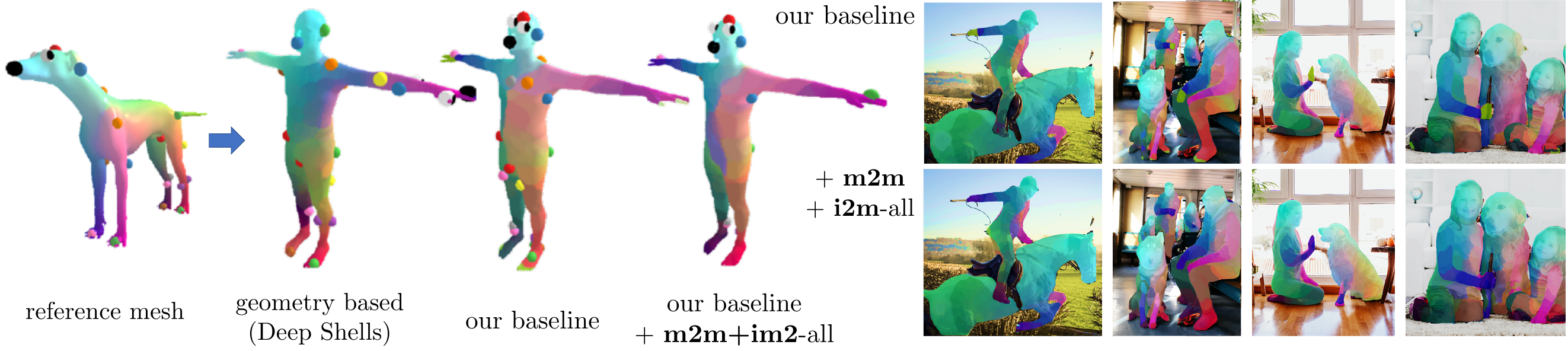}
\caption{\label{fig:3DalignSMPL} \textbf{3D model alignment: \texttt{dog}-\texttt{human}.} The \textbf{m2m+i2m}-all term is critical for aligning dissimilar categories in 3D.}
\end{figure*}

After discussing the experimental data and implementation details, we focus on our key contributions: simultaneously discovering high-quality inter-class correspondences while learning category-specific dense pose predictors (\cref{s:exp.3dalign,s:exp.canonicalmaps}).
We also show (\cref{s:exp.kptransfer}) that learned embeddings in the pixel space allow for effective retrieval of analogous points (body landmarks) on the surfaces of objects belonging to the same or different categories (a task that we call \textit{keypoint transfer}).

\paragraph{Training datasets.}


Following~\cite{neverova20continuous}, we use the original people-centric DensePose-COCO~\cite{guler18densepose:} for pre-training in all experiments. We also exploit this data for conducting studies on a joint set of animal and human categories.

For the animal classes, 
we propose an extended version of the DensePose-LVIS data of~\cite{neverova20continuous}, which originally contained 9 animal categories from the LVIS~v0.5 dataset for instance segmentation~\cite{gupta19lvis:}. 
Following a recent release of LVIS v1.0, which extended the benchmark to 160k images and 2M instance annotations, we also expand the DensePose annotation pool for the same animal classes and call this benchmark DensePose-LVIS~v1.0 
(see sup.~mat.).
The original DensePose-LVIS contains fairly sparse annotations (according to~\cite{neverova20continuous} only $18\%$ of the vertices of the animal meshes have at least one ground truth annotation, and each image contains no more than 3 annotated points).
While compared to DensePose-LVIS, we have 3.6$\times$ annotations, this is still far less than the original DensePose-COCO (which annotates 5 million points and obtained $96\%$ coverage of the SMPL mesh). At the same time, quality of dense labels in DensePose-LVIS v1.0 has been further improved by introducing an additional step of croud-sourced manual verification for all annotations.

\paragraph{Implementation details.}

Our architecture is similar to the R50 variant of~\cite{neverova20continuous}, with the only difference being the multi-class setting for object detection (in~\cite{neverova20continuous} detection was implemented in a class-agnostic manner, and ground truth class labels were required for inference).

We pre-train on the DensePose-COCO dataset for 130k iterations (following the standard s1x schedule).
All animal models are then trained on DensePose-LVIS v1.0 for 16k iterations, with a 10x drop of learning rate after 12k and 14k iterations, with the rest of hyperparameters being identical to~\cite{neverova20continuous}.
For experiments on the joint set of human and animal categories, we train our models for 80k iterations (with a learning rate decrease after 60k and 70k iterations).

\paragraph{Evaluation metrics.} 

The quality of learned \emph{dense pose predictions} is evaluated by a standard set of \textbf{AP}/\textbf{AR} metrics~\cite{guler18densepose:} (higher is better).
We also estimate the quality of \emph{inter-class mesh alignment} by computing the Geodesic Error ($\mathbf{GErr}$, lower is better) between the predicted and the ground truth vertices along the surface of the target mesh, given a set of manually annotated semantic keypoints.
For this purpose, all vertex coordinates in each mesh are normalised to have the maximum of geodesic distance $\mathbf{d}_{max}=2.27$ (analogously to~\cite{guler18densepose:,neverova20continuous}). Finally, we report Geodesic Point Similarity ($\mathbf{GPS}$, as in~\cite{guler18densepose:}, higher is better) as an alternative indicator of the quality of cross-category mesh alignment.


\subsection{Inter-class alignment and dense pose prediction}\label{s:exp.3dalign}


Compared to prior work such as~\cite{neverova20continuous}, our most important contribution is to discover automatically effective inter-category correspondences, without manual input for this task, while simultaneously learning high-quality canonical maps for each of the individual animal object categories.

In order to conduct a fair comparison, we start by rerunning the baseline of~\cite{neverova20continuous} using the embeddings and the DensePose-LVIS v1.0 data (\cref{tab:dp80k}).
We also compare using two different initializations for the embedding of the different 3D canonical shapes: random and ZoomOut.
The latter follows~\cite{neverova20continuous}, obtaining an initial set of inter-class 3D mesh correspondences from sparse manual annotations interpolated using the ZoomOut technique~\cite{melzi2019zoomout}.

We observe a 2.3pp AP gain in DensePose performance on the new dataset (AP $35.1\rightarrow 37.4$). While ZoomOut initialization of animal mesh embeddings provided a clear advantage for DensePose in a lower data regime (AP $25.4\rightarrow35.1$), the quality of mesh alignment worsens as the network diverges from its initialization point (GErr $23.7 \rightarrow 28.0$).
The dynamic on animal-only categories is similar. On DensePose-LVIS v1.0 the automatic alignment learned from random initialization is already of better quality than ZoomOut (20.7 GErr), and the difference in DensePose performance is no longer observed.
Note that the latter is already a confirmation of our key hypothesis that good inter-category correspondences should spontaneously emerge by jointly modelling them.

In~\cref{tab:dpresults} we report results on the animals-only benchmark, including assessing the contributions of the \textbf{m2m} and \textbf{i2m} regularisers.
Both \textbf{m2m} and \textbf{i2m}-all terms significantly improve mesh alignment (GErr $14.05 \rightarrow 12.67$, $11.74$, respectively) and also contribute to the dense pose performance (AP $37.5\rightarrow 38.1$, $38.3$, respectively). Their combination yeilds best results (GErr $14.05 \rightarrow 10.90$, AP $37.5\rightarrow 38.5$) and 
 fixes certain typical errors, as shown in~\cref{f:dpresults_i2m}. Fig.~\ref{f:dpresults} shows qualitative results.


\subsection{Further inter-class alignment analysis}\label{s:exp.canonicalmaps}

We compare the quality of inter-class mesh alignment produced by our networks with state-of-the-art methods exploiting 3D geometry: namely, ZoomOut~\cite{melzi2019zoomout} (initialized with the same manually keypoints, as we use for evaluation) and Deep Sheels~\cite{eisenberger2020deep} (unsupervised).
Qualitative and quantitative results on animal classes are shown in \cref{fig:3Dalign}.
Our method demonstrates consistently stronger performance across all categories, and rather successfully handles transfer between highly dissimilar categories, such as \texttt{dog-giraffe} and \texttt{dog-elephant}, where state-of-the-art geometry-based methods tend to fail (GErr ZoomOut: $26.24$, Deep Shells: $17.91$, and our best result: $9.42$).

An extreme case of \texttt{human-dog} alignment is shown in \cref{fig:3DalignSMPL}: our method produces meaningful correspondences in the 3D space (on the left) and consistent cross-category predictions in the  pixel space (on the bottom right, shown using the \textsc{dog} 3D mesh as a reference for visualization).

\begin{table*}[!t]
\centering
\scalebox{0.85}{%
\begin{tabular}{l|c|ccc|ccccc|c}
\toprule
\hspace*{\fill}\multirow{2}{*}{method}\hspace*{\fill} & \hspace*{\fill}\multirow{2}{*}{\parbox{1.2cm}{\centering target\\ class}}\hspace*{\fill}  & \multicolumn{3}{c|}{supervision} & \multicolumn{5}{c|}{animal category} & \multirow{2}{*}{mean} \\
\cmidrule{3-10}
&& mask\! & points\! & \!\!3D mesh & \textsc{horse}\! & \textsc{cow} & \!\textsc{sheep}\! & \textsc{cat} & \textsc{dog} & \\
\midrule
Rigid-CSM~\cite{kulkarni19canonical}  & single & \cmark & \xmark & \cmark & 31.2 & 26.3 & 24.7 & -- & -- & -- \\
Dense-Equi~\cite{thewlis17unsupervised} & single & \cmark & \xmark & \xmark & 23.3 & 20.9 & 19.6 & -- & -- & -- \\
A-CSM~\cite{kulkarni2020articulation} & single & \cmark & \xmark & \cmark & 32.9 & 26.3 & 28.6 & -- & -- & -- \\
\midrule
Rigid-CSM + keyp.~\cite{kulkarni19canonical}  & single & \cmark & \cmark* & \cmark & 42.1 & 28.5 & 31.5 & -- & -- & -- \\
A-CSM + keyp.~\cite{kulkarni2020articulation} & single & \cmark & \cmark* & \cmark & 44.6 & 29.2 & 39.0 & -- & -- & -- \\
\midrule
our baseline & multi &\cmark & \cmark & \xmark & 58.1 & 49.9 & 43.9 & 41.6 & 41.9 & 47.1 \\
w/ \textbf{m2m} & multi &\cmark & \cmark &  \xmark & 57.1 & 49.5 & 45.1 & 40.0 & 42.5 & 46.8 \\

w/ \textbf{i2m} & multi &\cmark & \cmark & \xmark & 59.0 & 51.1 & 46.2 & 45.9 & 45.7 & 49.7 \\
w/ \textbf{i2m}-all & multi &\cmark & \cmark & \xmark & \textbf{59.2} & \textbf{51.5} & \textbf{46.3} & \textbf{46.5} & \textbf{45.9} & \textbf{49.9} \\
\bottomrule
\end{tabular}
}
\caption{\label{tab:kps_transfer_within}\textbf{Keypoint transfer on PASCAL VOC, within each of training animal categories}. PCK-Transfer metric, higher is better. * -- supervision on the same set of keypoints that are used for evaluation, as opposed to random sampling in DensePose.}
\end{table*}

\begin{table}[!t]
\centering
\vspace{-10pt}  
\scalebox{0.82}{%
\begin{tabular}{cl|ccccc|l}
\toprule
&\hspace*{\fill}method\hspace*{\fill} & \!\textsc{horse}\!\! & \textsc{cow} & \!\!\textsc{sheep}\!\! & \textsc{cat} & \textsc{dog} & mean\\
\midrule
\multirow{3}{*}{(I)} & our baseline & 47.7 & 45.7 & 43.5 & 41.8 & 40.0 & 43.8 \\
& w/ \textbf{m2m} & 47.6 & 45.0 & 45.0 & 41.4 & 40.5 & 43.9 \\
& w/ \textbf{i2m} & \textbf{49.5} & \textbf{47.4} & \textbf{47.0} & \textbf{44.4} & \textbf{44.1} & \textbf{46.5} \\
\midrule
\multirow{2}{*}{(II)\!} & our baseline & 52.0 & 49.1 & 43.0 & 34.6 & 42.1 & 44.2\\
& w/ \textbf{i2m} & \textbf{54.6} & \textbf{49.5} & \textbf{44.7} & \textbf{37.7} & \textbf{43.7} & \textbf{46.0}\\
\bottomrule
\end{tabular}
} \vspace*{-2pt}%
\caption{\label{tab:kps_transfer}\textbf{Keypoint transfer on PASCAL VOC:} (I) across training categories, (II) within new animal categories not observed during training for dense correspondences (only boxes and masks are provided during training to ensure robust  detection). PCK-Transfer metric, higher is better.}\vspace*{-6pt}
\end{table}

\begin{figure}[!t]
\begin{center}
\includegraphics[width=0.89\linewidth]{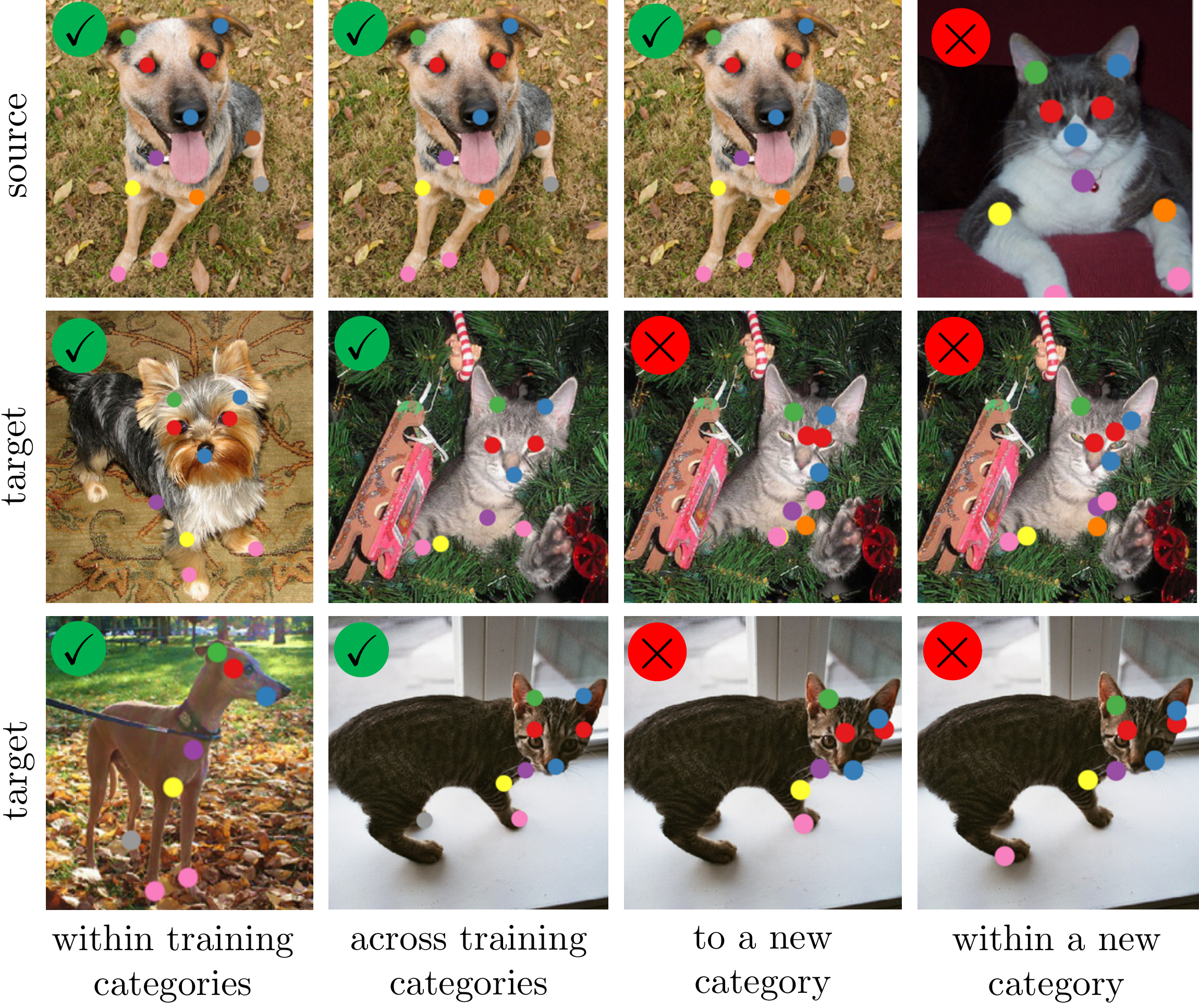}\vspace*{-5pt}
\caption{\textbf{Keypoint transfer on PASCAL VOC.} One experiment -- one column. Green marks indicate categories included in training, red marks -- a new, test only category.\vspace*{-15pt}}\label{f:kpresults}
\end{center}
\end{figure}

\subsection{Keypoint transfer}\label{s:exp.kptransfer}

To evaluate learned transferability of surface embeddings within and across training categories, as well as their ability to generalize to new animal classes, we look at the problem of \textit{keypoint transfer}.
As per~\cref{s:cse}, we can in fact use  our learned embeddings to establish correspondences between a source image $I$ and a target image $I'$ directly and use it to transfer keypoints.
To do this, in the target image, for each type of landmark annotated as ``visible'', we take the nearest neighbor of the embedding of the pixel in the source image, corresponding to the same landmark.

\paragraph{Evaluation metric.}

Following~\cite{kulkarni2020articulation}, we evaluate performance on this task by estimating the Percentage of Correct Keypoint transfers (PCK-Transfer). The transfer is said to be correct if the target landmark is localized within distance $0.1\cdot\max(h,w)$ from the annotated location, where $h,w$ are the height and the width of the predicted bounding box.
Prior to that, we match predicted object instances to ground truth objects by estimating the bounding box IoU. Objects that are not retrieved are excluded from evaluation. We report performance on animal categories from PASCAL VOC~\cite{everingham15the-pascal}, overlapping with animal categories in DensePose-LVIS v1.0 (\texttt{horse}, \texttt{cow}, \texttt{sheep}, \texttt{cat}, \texttt{dog}).

\paragraph{Experimental protocol.} We evaluate the ability of our model to perform keypoint transfer in three distinct settings: (a) within each category observed at training time (Tab.~\ref{tab:kps_transfer_within}); (b) across training categories (Tab.~\ref{tab:kps_transfer}, I);
(c) within new animal categories (zero shot) not observed at training time.
For (c), dense correspondences for one class are removed from the training set, and only bounding boxes and object instance masks are provided as supervision (Tab.~\ref{tab:kps_transfer}, II).

\paragraph{Discussion.}

As shown in~\cref{tab:kps_transfer_within,tab:kps_transfer,f:kpresults}, the learned embeddings work very well to transfer keypoints between known as well as unknown animal classes, demonstrating once more the power of generalization that comes from joint modelling. 
For this experiment, the \textbf{i2m} regularization term significantly improves the results (\eg \textbf{m2m} vs \textbf{i2m} PCK\@: $46.8\rightarrow 49.9$ in~\cref{tab:kps_transfer}).
This might be expected since \textbf{m2m} works on the alignment between 3D templates, whereas \textbf{i2m} works at the image level, which is more relevant for this experiment.
Note that methods~\cite{kulkarni19canonical,kulkarni2020articulation} re-project a 3D mesh template onto the images to establish correspondences, which we do not do.
\section{Conclusions}\label{s:conclusions}

We have introduced a method to learn high-quality dense pose predictors for multiple object categories while discovering automatically semantic correspondences between them.
The method represents correspondences via a unified embedding and network predictor while enforcing reasonable topological consistency constraints.
Our encouraging results indicate that joint modelling is not only just viable, but has significant positive effects on performance and scalability of such neural dense predictors.

{\small\bibliographystyle{ieee_fullname}\bibliography{refs,vedaldi_general,vedaldi_specific}}

\begin{thebibliography}{10}\itemsep=-1pt

\bibitem{andriluka142d-human}
Mykhaylo Andriluka, Leonid Pishchulin, Peter~V. Gehler, and Bernt Schiele.
\newblock {2D} human pose estimation: New benchmark and state of the art
  analysis.
\newblock In {\em Proc. {CVPR}}, 2014.

\bibitem{aubry2011wave}
Mathieu Aubry, Ulrich Schlickewei, and Daniel Cremers.
\newblock The wave kernel signature: A quantum mechanical approach to shape
  analysis.
\newblock In {\em IEEE International Conference on Computer Vision Workshops
  (ICCV Workshops)}, pages 1626--1633, 2011.

\bibitem{biggs2020left}
Benjamin Biggs, Oliver Boyne, James Charles, Andrew Fitzgibbon, and Roberto
  Cipolla.
\newblock Who left the dogs out? 3d animal reconstruction with expectation
  maximization in the loop.
\newblock {\em arXiv preprint arXiv:2007.11110}, 2020.

\bibitem{biggs2018}
Benjamin Biggs, Thomas Roddick, Andrew Fitzgibbon, and Roberto Cipolla.
\newblock {C}reatures {G}reat and {SMAL}: {R}ecovering the {S}hape and {M}otion
  of {A}nimals from {V}ideo.
\newblock In {\em Asian Conference on Computer Vision (ACCV)}, pages 3--19,
  2018.

\bibitem{bourdev09poselets:}
Lubomir~D. Bourdev and Jitendra Malik.
\newblock Poselets: Body part detectors trained using {3D} human pose
  annotations.
\newblock In {\em Proc. {ICCV}}, 2009.

\bibitem{bronstein2006generalized}
Alexander~M. Bronstein, Michael~M. Bronstein, and Ron Kimmel.
\newblock Generalized multidimensional scaling: a framework for
  isometry-invariant partial surface matching.
\newblock {\em Proceedings of the National Academy of Sciences (PNAS)},
  103(5):1168--1172, 2006.

\bibitem{bronstein2010gromov}
Alexander~M. Bronstein, Michael~M. Bronstein, Ron Kimmel, Mona Mahmoudi, and
  Guillermo Sapiro.
\newblock A {G}romov-{H}ausdorff framework with {D}iffusion {G}eometry for
  {T}opologically-{R}obust {N}on-rigid {S}hape {M}atching.
\newblock {\em International Journal of Computer Vision}, 89(2--3):266--286,
  2010.

\bibitem{bronstein2010scale}
Michael~M. Bronstein and Iasonas Kokkinos.
\newblock {S}cale-invariant heat kernel signatures for non-rigid shape
  recognition.
\newblock In {\em IEEE Conference on Computer Vision and Pattern Recognition
  (CVPR)}, pages 1704 -- 1711, 2010.

\bibitem{burgin18how-many}
Connor~J Burgin, Jocelyn~P Colella, Philip~L Kahn, and Nathan~S Upham.
\newblock {How many species of mammals are there?}
\newblock {\em Journal of Mammalogy}, 99(1):1--14, 02 2018.

\bibitem{cao17realtime}
Zhe Cao, Tomas Simon, Shih{-}En Wei, and Yaser Sheikh.
\newblock Realtime multi-person {2D} pose estimation using part affinity
  fields.
\newblock In {\em Proc. {CVPR}}, 2017.

\bibitem{coifman2005diffusion}
Ronald~R. Coifman and Stéphane Lafon.
\newblock Diffusion maps.
\newblock {\em Applied and Computational Harmonic Analysis}, 21(1):5--30, 2006.

\bibitem{eisenberger2020deep}
Marvin Eisenberger, Aysim Toker, Laura Leal-Taixe, and Daniel Cremers.
\newblock Deep shells: Unsupervised shape correspondence with optimal
  transport.
\newblock {\em arXiv preprint}, 2020.

\bibitem{elad2003bending}
Asi~Elad (Elbaz) and Ron Kimmel.
\newblock {O}n {B}ending {I}nvariant {S}ignatures for {S}urfaces.
\newblock {\em IEEE Transactions on Pattern Analysis and Machine Intelligence},
  25(10):1285--1295, 2003.

\bibitem{everingham15the-pascal}
Mark Everingham, S.~M.~Ali Eslami, Luc~Van Gool, Christopher K.~I. Williams,
  John~M. Winn, and Andrew Zisserman.
\newblock The pascal visual object classes challenge: {A} retrospective.
\newblock {\em {IJCV}}, 111(1), 2015.

\bibitem{felzenszwalb08a-discriminatively}
Pedro~F. Felzenszwalb, David~A. McAllester, and Deva Ramanan.
\newblock A discriminatively trained, multiscale, deformable part model.
\newblock In {\em Proc. {CVPR}}, 2008.

\bibitem{girshick13rich}
R.~B. Girshick, J. Donahue, T. Darrell, and J. Malik.
\newblock Rich feature hierarchies for accurate object detection and semantic
  segmentation.
\newblock {\em arXiv preprint arXiv:1311.2524}, 2013.

\bibitem{guler18densepose:}
Riza~Alp G{\"{u}}ler, Natalia Neverova, and Iasonas Kokkinos.
\newblock {DensePose}: Dense human pose estimation in the wild.
\newblock In {\em Proc. {CVPR}}, 2018.

\bibitem{gupta19lvis:}
Agrim Gupta, Piotr Doll{\'{a}}r, and Ross~B. Girshick.
\newblock {LVIS:} {A} dataset for large vocabulary instance segmentation.
\newblock In {\em Proc. {CVPR}}, 2019.

\bibitem{huang13consistent}
Qi{-}Xing Huang and Leonidas~J. Guibas.
\newblock Consistent shape maps via semidefinite programming.
\newblock {\em Eurographics Symposium on Geometry Processing}, 32(5), 2013.

\bibitem{huang2013consistent}
Qi-Xing Huang and Leonidas Guibas.
\newblock Consistent shape maps via semidefinite programming.
\newblock In {\em Computer Graphics Forum}, volume~32, pages 177--186. Wiley
  Online Library, 2013.

\bibitem{huber02automatic}
Daniel Huber.
\newblock {\em Automatic Three-dimensional Modeling from Reality}.
\newblock PhD thesis, Carnegie Mellon University, 2002.

\bibitem{johnson10clustered}
Sam Johnson and Mark Everingham.
\newblock Clustered pose and nonlinear appearance models for human pose
  estimation.
\newblock In {\em BMVC}, 2010.

\bibitem{johnson11learning}
Sam Johnson and Mark Everingham.
\newblock Learning effective human pose estimation from inaccurate annotation.
\newblock In {\em Proc. {CVPR}}, 2011.

\bibitem{kanazawa18end-to-end}
Angjoo Kanazawa, Michael~J. Black, David~W. Jacobs, and Jitendra Malik.
\newblock End-to-end recovery of human shape and pose.
\newblock In {\em Proc. {CVPR}}, 2018.

\bibitem{kanazawa16warpnet:}
Angjoo Kanazawa, David~W. Jacobs, and Manmohan Chandraker.
\newblock {WarpNet}: Weakly supervised matching for single-view reconstruction.
\newblock In {\em Proc. {CVPR}}, 2016.

\bibitem{kanazawa18learning}
Angjoo Kanazawa, Shubham Tulsiani, Alexei~A. Efros, and Jitendra Malik.
\newblock Learning category-specific mesh reconstruction from image
  collections.
\newblock In {\em Proc. {ECCV}}, 2018.

\bibitem{kolotouros19learning}
Nikos Kolotouros, Georgios Pavlakos, Michael~J Black, and Kostas Daniilidis.
\newblock Learning to reconstruct {3D} human pose and shape via model-fitting
  in the loop.
\newblock In {\em ICCV}, 2019.

\bibitem{kolotouros19convolutional}
Nikos Kolotouros, Georgios Pavlakos, and Kostas Daniilidis.
\newblock Convolutional mesh regression for single-image human shape
  reconstruction.
\newblock In {\em Proc. {CVPR}}, 2019.

\bibitem{kulkarni2020articulation}
Nilesh Kulkarni, Abhinav Gupta, David~F Fouhey, and Shubham Tulsiani.
\newblock Articulation-aware canonical surface mapping.
\newblock In {\em Proceedings of the IEEE/CVF Conference on Computer Vision and
  Pattern Recognition}, pages 452--461, 2020.

\bibitem{kulkarni19canonical}
Nilesh Kulkarni, Abhinav Gupta, and Shubham Tulsiani.
\newblock Canonical surface mapping via geometric cycle consistency.
\newblock In {\em Proc. {ICCV}}, 2019.

\bibitem{lin14microsoft}
Tsung{-}Yi Lin, Michael Maire, Serge~J. Belongie, James Hays, Pietro Perona,
  Deva Ramanan, Piotr Doll{\'{a}}r, and C.~Lawrence Zitnick.
\newblock Microsoft {COCO:} common objects in context.
\newblock In {\em Proc. {ECCV}}, 2014.

\bibitem{loper15smpl:}
Matthew Loper, Naureen Mahmood, Javier Romero, Gerard Pons-Moll, and Michael~J.
  Black.
\newblock {SMPL}: a skinned multi-person linear model.
\newblock {\em ACM Trans. on Graphics (TOG)}, 2015.

\bibitem{melzi2019zoomout}
Simone Melzi, Jing Ren, Emanuele Rodol{\`a}, Abhishek Sharma, Peter Wonka, and
  Maks Ovsjanikov.
\newblock Zoomout: Spectral upsampling for efficient shape correspondence.
\newblock {\em ACM Transactions on Graphics (TOG)}, 38(6):155, 2019.

\bibitem{mora11how-many}
Camilo Mora, Derek~P. Tittensor, Sina Adl, Alastair G.~B. Simpson, and Boris
  Worm.
\newblock How many species are there on earth and in the ocean?
\newblock {\em PLOS Biology}, 9(8), 2011.

\bibitem{neverova20continuous}
Natalia Neverova, David Novotn{\'{y}}, and Andrea Vedaldi.
\newblock Continuous surface embeddings.
\newblock In {\em Proceedings of Advances in Neural Information Processing
  Systems (NeurIPS)}, 2020.

\bibitem{neverova19slim}
Natalia Neverova, James Thewlis, Riza~Alp G{\"{u}}ler, Iasonas Kokkinos, and
  Andrea Vedaldi.
\newblock Slim {DensePose}: Thrifty learning from sparse annotations and motion
  cues.
\newblock In {\em Proceedings of the {IEEE} Conference on Computer Vision and
  Pattern Recognition ({CVPR})}, 2019.

\bibitem{newell16stacked}
Alejandro Newell, Kaiyu Yang, and Jia Deng.
\newblock Stacked hourglass networks for human pose estimation.
\newblock In {\em Proc. {ECCV}}, 2016.

\bibitem{novotny19c3dpo}
David Novotn{\'{y}}, Nikhila Ravi, Benjamin Graham, Natalia Neverova, and
  Andrea Vedaldi.
\newblock {C3DPO}: Canonical 3d pose networks for non-rigid structure from
  motion.
\newblock In {\em Proceedings of the International Conference on Computer
  Vision ({ICCV})}, 2019.

\bibitem{ovsjanikov12functional}
Maks Ovsjanikov, Mirela Ben{-}Chen, Justin Solomon, Adrian Butscher, and
  Leonidas~J. Guibas.
\newblock Functional maps: a flexible representation of maps between shapes.
\newblock {\em {ACM} Trans. Graph.}, 31(4), 2012.

\bibitem{ramanan06learning}
Deva Ramanan.
\newblock Learning to parse images of articulated bodies.
\newblock In {\em Proc. {NeurIPS}}, 2006.

\bibitem{ren2020maptree}
Jing Ren, Simone Melzi, Maks Ovsjanikov, and Peter Wonka.
\newblock Maptree: recovering multiple solutions in the space of maps.
\newblock {\em ACM Transactions on Graphics (TOG)}, 39(6):1--17, 2020.

\bibitem{rustamov2007laplace}
Raif~M. Rustamov.
\newblock Laplace-{B}eltrami eigenfunctions for deformation invariant shape
  representation.
\newblock In {\em Symposium on Geometry Processing}, pages 225--233, 2007.

\bibitem{sanakoyeu20transferring}
Artsiom Sanakoyeu, Vasil Khalidov, Maureen~S. McCarthy, Andrea Vedaldi, and
  Natalia Neverova.
\newblock Transferring dense pose to proximal animal classes.
\newblock In {\em Proceedings of the {IEEE} Conference on Computer Vision and
  Pattern Recognition ({CVPR})}, 2020.

\bibitem{singh2016learning}
Saurabh Singh, Derek Hoiem, and David~A. Forsyth.
\newblock {L}earning to {L}ocalize {L}ittle {L}andmarks.
\newblock In {\em IEEE Conference on Computer Vision and Pattern Recognition
  (CVPR)}, pages 260--269, 2016.

\bibitem{sun2009concise}
Jian Sun, Maks Ovsjanikov, and Leonidas~J. Guibas.
\newblock {A} {C}oncise and {P}rovably {I}nformative {M}ulti-{S}cale
  {S}ignature {B}ased on {H}eat {D}iffusion.
\newblock {\em Computer Graphics Forum}, 28(5):1383--1392, 2009.

\bibitem{thewlis17unsupervised}
James Thewlis, Hakan Bilen, and Andrea Vedaldi.
\newblock Unsupervised learning of object landmarks by factorized spatial
  embeddings.
\newblock In {\em Proceedings of the International Conference on Computer
  Vision ({ICCV})}, 2017.

\bibitem{wang19learning}
Xiaolong Wang, Allan Jabri, and Alexei~A. Efros.
\newblock Learning correspondence from the cycle-consistency of time.
\newblock In {\em Proc. {CVPR}}, 2019.

\bibitem{wei16convolutional}
Shih{-}En Wei, Varun Ramakrishna, Takeo Kanade, and Yaser Sheikh.
\newblock Convolutional pose machines.
\newblock In {\em Proc. {CVPR}}, 2016.

\bibitem{welinder10caltech-ucsd}
P. Welinder, S. Branson, T. Mita, C. Wah, and F. Schroff.
\newblock Caltech-ucsd birds 200.
\newblock Technical report, 2010.

\bibitem{yang20mapping}
Lei Yang, Wenxi Liu, Zhiming Cui, Nenglun Chen, and Wenping Wang.
\newblock Mapping in a cycle: Sinkhorn regularized unsupervised learning for
  point cloud shapes.
\newblock In {\em Proc. {ECCV}}, 2020.

\bibitem{yang2019parsing}
Lu Yang, Qing Song, Zhihui Wang, and Ming Jiang.
\newblock Parsing r-cnn for instance-level human analysis.
\newblock In {\em Proceedings of the IEEE Conference on Computer Vision and
  Pattern Recognition}, pages 364--373, 2019.

\bibitem{zhang14part-based}
Ning Zhang, Jeff Donahue, Ross~B. Girshick, and Trevor Darrell.
\newblock Part-based {R-CNNs} for fine-grained category detection.
\newblock In {\em Proc. {ECCV}}, 2014.

\bibitem{zhou2016learning}
Tinghui Zhou, Philipp Krahenbuhl, Mathieu Aubry, Qixing Huang, and Alexei~A
  Efros.
\newblock Learning dense correspondence via 3d-guided cycle consistency.
\newblock In {\em Proceedings of the IEEE Conference on Computer Vision and
  Pattern Recognition}, pages 117--126, 2016.

\bibitem{zhou16blearning}
Tinghui Zhou, Philipp Kr{\"{a}}henb{\"{u}}hl, Mathieu Aubry, Qi{-}Xing Huang,
  and Alexei~A. Efros.
\newblock Learning dense correspondence via {3D-Guided} cycle consistency.
\newblock In {\em Proc. {CVPR}}, 2016.

\bibitem{zhou15flowweb:}
Tinghui Zhou, Yong~Jae Lee, Stella~X. Yu, and Alexei~A. Efros.
\newblock {FlowWeb}: Joint image set alignment by weaving consistent,
  pixel-wise correspondences.
\newblock In {\em Proc. {CVPR}}, 2015.

\bibitem{zhou2015multi}
Xiaowei Zhou, Menglong Zhu, and Kostas Daniilidis.
\newblock Multi-image matching via fast alternating minimization.
\newblock In {\em Proceedings of the IEEE International Conference on Computer
  Vision}, pages 4032--4040, 2015.

\bibitem{zhu17unpaired}
Jun{-}Yan Zhu, Taesung Park, Phillip Isola, and Alexei~A. Efros.
\newblock Unpaired image-to-image translation using cycle-consistent
  adversarial networks.
\newblock In {\em Proc. {ICCV}}, 2017.

\bibitem{zuffi2019safari}
Silvia Zuffi, Angjoo Kanazawa, Tanya~Y. Berger{-}Wolf, and Michael~J. Black.
\newblock {T}hree-{D} {S}afari: {L}earning to {E}stimate {Z}ebra {P}ose,
  {S}hape, and {T}exture from {I}mages "{I}n the {W}ild".
\newblock In {\em International Conference on Computer Vision (ICCV)}, pages
  5358--5367, 2019.

\bibitem{zuffi2017menagerie}
Silvia Zuffi, Angjoo Kanazawa, David~W. Jacobs, and Michael~J. Black.
\newblock {3D} {M}enagerie: {M}odeling the {3D} {S}hape and {P}ose of
  {A}nimals.
\newblock In {\em IEEE Conference on Computer Vision and Pattern Recognition
  (CVPR)}, pages 5524--5532, 2017.

\bibitem{zuffi2018lions}
Silvia Zuffi, Angjoo Kanazawa, David~W. Jacobs, and Michael~J. Black:.
\newblock {L}ions and {T}igers and {B}ears: {C}apturing {N}on-{R}igid, {3D},
  {A}rticulated {S}hape from {I}mages.
\newblock In {\em IEEE Conference on Computer Vision and Pattern Recognition
  (CVPR)}, pages 3955--3963, 2018.

\end{thebibliography}

\clearpage
\appendix
\noindent{\LARGE \textbf{Appendix}}
\vspace{12pt}


Project page: \href{https://gdude.de/discovering-3d-obj-rel}{https://gdude.de/discovering-3d-obj-rel}.

\section{Experiments}
\subsection{DensePose-LVIS~v1.0 dataset details}
We introduce DensePose-LVIS~v1.0 dataset, an extended version of the DensePose-LVIS data of~\cite{neverova20continuous}. We improve the quality of the existing labels and expand the DensePose annotation pool for the same animal classes as in the previous version of this dataset~\cite{neverova20continuous}. In Tab.~\ref{tab:lvispp} we report the number of train and test instances annotated with DensePose in every category.

\begin{table}[!h]
\centering
\scalebox{0.8}{
\begin{tabular}{l|cc|cc}
\toprule
\hspace*{\fill}\multirow{2}{*}{category}\hspace*{\fill} & \multicolumn{2}{c|}{DensePose-LVIS} & \multicolumn{2}{c}{DensePose-LVIS v1.0}\\
\cmidrule{2-5}
& train, inst. & test, inst. & train, inst. & test, inst. \\
  \midrule
\texttt{dog} & 483 & 200 & 1607 & 316 \\
\texttt{cat} &  586 & 200 & 1912 & 379 \\
\texttt{bear} &  98 & 200 & 735 & 132 \\
\texttt{sheep} &  257 & 200 & 1655 & 350 \\
\texttt{cow} &  426 & 200 & 2105 & 340\\
\texttt{horse} &  605 & 200 & 2292 & 458\\
\texttt{zebra} &  665 & 200 & 2864 & 556\\
\texttt{giraffe} &  651 & 200 & 2709 & 534\\
\texttt{elephant} &  670 & 200 & 2839 & 539\\
\midrule
all & 4441 & 1800 & 18718 & 3604\\
\bottomrule
\end{tabular}}
\caption{\textbf{DensePose-LVIS v1.0 dataset:}  3.6x increase in a number of annotated instances, better quality of labels.}
\label{tab:lvispp}
\end{table}

\subsection{3D mesh alignment}
If not stated otherwise, we used cross-validation to find the \textbf{m2m} loss weight that maximizes $\textbf{AP}$ metric after training. Additionally we also cross-validated the \textbf{m2m} loss weight to minimize the $\textbf{GErr}$, we denote this experiment as \textbf{m2m*}. As mentioned in the caption of Fig.~4, the optimal weight of \textbf{m2m} term is tenfold larger for $\textbf{GErr}$ than for $\textbf{AP}$. 
Visual mappings in Fig.~4 correspond to the \textbf{m2m} model.

\subsection{Keypoint transfer}
For keypoint transfer experiments we train our models on DensePose-LVIS v1.0 dataset and do not use any PASCAL VOC~\cite{everingham15the-pascal} images during training.
We select animal categories from PASCAL VOC~\cite{everingham15the-pascal}, overlapping with animal categories in DensePose-LVIS v1.0: \texttt{horse}, \texttt{cow}, \texttt{sheep}, \texttt{cat}, \texttt{dog}. Following Kulkarni et al.~\cite{kulkarni2020articulation}, we randomly sample $100$  images for each category from PASCAL VOC mentioned above and use them for evaluation. We average PCK-Transfer score across all possible (source, target) image pairs.

\begin{table}[!t]
\centering
\scalebox{0.8}{%
\begin{tabular}{l|ccccc|c}
\toprule
\hspace*{\fill}\multirow{2}{*}{method}\hspace*{\fill} & \multicolumn{5}{c|}{animal category} & \multirow{2}{*}{mean} \\
\cmidrule{2-6}
& \textsc{horse}\! & \textsc{cow} & \!\textsc{sheep}\! & \textsc{cat} & \textsc{dog} & \\
\midrule
our baseline & 58.1 & 49.9 & 43.9 & 41.6 & 41.9 & 47.1 \\
w/ \textbf{m2m} & 57.1 & 49.5 & 45.1 & 40.0 & 42.5 & 46.8 \\
w/ \textbf{i2m} &  59.0 & 51.1 & 46.2 & 45.9 & 45.7 & 49.7 \\
w/ \textbf{i2m}-all & \textbf{59.2} & \textbf{51.5} & \textbf{46.3} & \textbf{46.5} & \textbf{45.9} & \textbf{49.9} \\

w/ \textbf{m2m+i2m} &  57.7 & 49.9 & 44.8 & 40.6 & 42.4 & 47.1 \\
w/ \textbf{m2m+i2m}-all &  57.8 & 50.2 & 44.9 & 40.6 & 42.6 & 47.2 \\
\bottomrule
\end{tabular}
}
\caption{\label{tab:kps_transfer_within_full}\textbf{Keypoint transfer on PASCAL VOC, within each of training animal categories}. PCK-Transfer metric, higher is better. \textbf{m2m} term is not helpful for this task.}
\end{table}

We conducted keypoint transfer experiments using three distinct settings: 
\begin{itemize}[(a)]
    \item Within each category observed at training time, when source and target images are from the same category (Tab.~3 in the main paper); 
    \item Across training categories (Tab.~4, I in the main paper). In this case source and target images are from different training categories. For example, keypoints from \texttt{dog} images are transferred to images from \texttt{horse}, \texttt{cow}, \texttt{sheep}, and \texttt{cat} categories; 
    \item Zero shot scenario: Within new animal categories not observed at training time (Tab.~4 II in the main paper).
In this case, we remove ground truth dense correspondences for one class from the training set and evaluate keypoint transfer on the images within the removed class. Note that we do not remove bounding boxes and object instance masks from training set and still use them to train our detection and segmentation heads.

\end{itemize}

\begin{table*}[!t]
\centering
\scalebox{0.9}{
\begin{tabular}{l|c|ccccc|l}
\toprule
method\hspace*{\fill} & target class & \textsc{horse} & \textsc{cow} & \textsc{sheep} & \textsc{cat} & \textsc{dog} & mean\\
\midrule
our baseline & single & 28.8 & 27.3 & 33.7 & 31.1 & 29.4 & 30.0 \\
w/ \textbf{i2m} & single & 30.1 & 28.2 & 34.1 & 31.9 & 29.7 & 30.8\\
\bottomrule
\end{tabular}
}
\caption{\textbf{Effect of \textbf{i2m} loss term when trained and evaluated on individual categories of  DensePose-LVIS v1.0.} We report DensePose \textbf{AP} score.}
\label{tab:dp_lvis_ap_single_class_train}
\end{table*}

\begin{table*}[!t]
\centering
\scalebox{0.9}{
\begin{tabular}{l|c|ccccc|l}
\toprule
method\hspace*{\fill} & target class & \textsc{horse} & \textsc{cow} & \textsc{sheep} & \textsc{cat} & \textsc{dog} & mean\\
\midrule
Rigid-CSM + keyp.~\cite{kulkarni19canonical}  & single & 42.1 & 28.5 & 31.5 & -- & -- & -- \\
A-CSM + keyp.~\cite{kulkarni2020articulation} & single & 44.6 & 29.2 & 39.0 & -- & -- & -- \\
\midrule
our baseline & single & 53.4 &  48.0 & 38.8 & 40.9 & 34.0 & 43.0 \\
w/ \textbf{i2m} & single & 54.0 & 49.1 & 39.2 & 34.7 & 44.2 & 44.2 \\
\bottomrule
\end{tabular}
}
\caption{\textbf{Keypoint transfer on PASCAL VOC, when trained and evaluated on individual categories.} PCK-Transfer metric, higher is better.}
\label{tab:kps_transfer_single_class_train}
\end{table*}


In Tab.~\ref{tab:kps_transfer_within_full} we show results for all combinations of the loss terms on keypoint transfer within each category.  The \textbf{i2m} term enforces alignment on image level which is more important for the task of within-category keypoint transfer, while the \textbf{m2m} loss term is not helpful in this case as it enforces the alignment between 3D templates.

\subsection{Effect of i2m loss in a single-class training scenario}
In contrast to our approach, Rigid-CSM~\cite{kulkarni19canonical} and A-CSM~\cite{kulkarni2020articulation} cannot learn multiple animal categories in a single model and have to train separate models for every animal class. To make our setup closer to those in \cite{kulkarni19canonical, kulkarni2020articulation}, we trained our models on individual categories from DensePose-LVIS v1.0 as well (i.e., trained a new model for each class). Then we evaluated each model on the corresponding individual categories: (a) on the test set of DensePose-LVIS v1.0 by computing DensePose AP score (see Tab.~\ref{tab:dp_lvis_ap_single_class_train}); and (b) on PASCAL VOC  by computing PCK-Transfer metric for Keypoint Transfer task (see Tab.~\ref{tab:kps_transfer_single_class_train}). From the tables~\ref{tab:dp_lvis_ap_single_class_train}, \ref{tab:kps_transfer_single_class_train} we can see that  \textbf{i2m} improves the performance even when trained and evaluated on individual categories and outperforms methods \cite{kulkarni19canonical, kulkarni2020articulation} (we used the results reported in the corresponding papers). However, a part of the strength of our method comes from training on several classes jointly, which results in even stronger performance of our models (see Tab.~3 in the main paper).

\subsubsection{Can the \textbf{i2m} loss be used in combination with Rigid-CSM~\cite{kulkarni19canonical} or A-CSM~\cite{kulkarni2020articulation} models?}

Our \textbf{i2m} loss requires every pixel and every vertex of the mesh to be embedded in a common embedding space. However, models \cite{kulkarni19canonical,kulkarni2020articulation} directly predict $(u,v)$ coordinates for every pixel. Therefore our loss cannot be applied during training of Rigid-CSM~\cite{kulkarni19canonical} and A-CSM~\cite{kulkarni2020articulation}.

\subsection{Evaluation metrics}
For completeness, we provide brief descriptions of $\textbf{AP}$/$\textbf{AR}$ metrics used for evaluation of learned dense pose predictions on DensePose-LVIS v1.0 dataset and cross-category mesh alignment metrics $\mathbf{GErr}$, $\mathbf{GPS}$.

\begin{itemize}
    \item $\mathbf{GPS}$ (Geodesic Point Similarity) \cite{guler18densepose:} is a correspondence matching score indicator of the quality of aligning of two sets of vertices $A=(a_1, \dots, a_N)$ and $B=(b_1, \dots, a_N)$ on a mesh.
    
    $$\mathbf{GPS}(A, B) = \frac{1}{N} \sum \limits_{i = 1}^N \exp \left( \frac{-g(a_i, b_i)^2}{2\kappa^2} \right),$$
    where $N$ is the number of vertices in each set, $g(\cdot, \cdot)$ is the geodesic distance between two surface points, and $\kappa$ is a normalization constant. 
    To make our $\mathbf{GPS}$ score comparable to the $\mathbf{GPS}$ score used for Human DensePose evaluation in Guler et al.~\cite{guler18densepose:}, we normalize all vertex coordinates in every animal mesh to have the maximum geodesic distance $d_{max}= 2.27$, which is equal to the maximal geodesic distance in the SMPL~\cite{loper15smpl:} mesh of a human utilized in~\cite{guler18densepose:}. We set $\kappa=0.255$ so that a single point has a $\mathbf{GPS}$ value of $0.5$ if its geodesic distance from the ground truth equals the average half-size of a body segment.
    When we evaluate cross-category mesh alignment quality, we compute $\mathbf{GPS}$ between a set of the ground truth semantic keypoints on a target mesh and the estimated locations of these keypoints obtained by transferring keypoints from a source mesh of other category.
    The mean $\mathbf{GPS}$ score is then computed as an average across all possible (source, target) pairs of categories.
    \item Similarly to Geodesic Point Similarity, we define $\mathbf{GErr}$ (Geodesic Error), the error between  two sets of vertices along the surface of a mesh. Before computing this error, all vertex coordinates are normalised to have the maximum of geodesic distance $d_{max} = 2.27$ (similar to \cite{guler18densepose:,neverova20continuous}). 
     $$\mathbf{GErr}(A, B) = \frac{1}{|N|} \sum  \limits_{i = 1}^N g(a_i, b_i).$$
    Analogously to $\mathbf{GPS}$, we use $\mathbf{GErr}$ to estimate the quality of inter-category mesh alignment by comparing the ground truth semantic keypoints on a target mesh and the estimated locations of these keypoints obtained by transferring keypoints from a source mesh of other category.
    
    \item  To evaluate the quality of mapping from image pixels to 3D vertices on the category-specific mesh, we use $\textbf{AP}$ (Average Precision) and $\textbf{AR}$ (Average Recall) \cite{guler18densepose:}. The location of the vertices on the mesh corresponding to image pixels are estimated by finding for every pixel the most similar mesh vertex in the learned embedding space. After that we compare estimated vertex locations with the ground truth using $\mathbf{GPS}$ metric.
    Then we calculate $\textbf{AP}$ and $\textbf{AR}$ at different $\mathbf{GPS}$ thresholds ranging from $0.5$ to $0.95$, following the COCO challenge protocol \cite{lin14microsoft}. We separately report Average Precision and Average Recall at $\mathbf{GPS}$ thresholds equal to $0.5$ and $0.75$, denoted as $\textbf{AP}_{50}, \textbf{AP}_{75}, \textbf{AR}_{50}, \textbf{AR}_{75}$. In addition to this we separately compute Average Precision and Average Recall for instances with \emph{medium} and \emph{large} sizes ($\textbf{AP}_{M}, \textbf{AR}_{M}$ for medium size and $\textbf{AP}_{L}, \textbf{AR}_{L}$ for large). 
\end{itemize}

Note, that we report $\mathbf{GPS}\times100$ and $\mathbf{GErr}\times100$ in all tables in the main paper.

\end{document}